\documentclass[letterpaper]{article}
\usepackage{aaai}
\usepackage{times}
\usepackage{helvet}
\usepackage{courier}
\usepackage{graphicx}
\usepackage{amssymb}
\usepackage{cite}
\begin{document}
\title{A Deep Learning Approach for Expert Identification \\
in Question Answering Communities}
\author{Chen Zheng\\Computer Science Department \\Binghamton University\\ Binghamton, NY 13902,USA\\czheng17@binghamton.edu
\And
Shuangfei Zhai \\Computer Science Department\\Binghamton University \\ Binghamton, NY 13902,USA\\szhai2@binghamton.edu \\
\And 
 Zhongfei Zhang\\Computer Science Department\\Binghamton University \\ Binghamton, NY 13902,USA\\zhongfei@cs.binghamton.edu\\
}
\maketitle
\begin{abstract}
\begin{quote}
In this paper, we describe an effective convolutional neural network framework for identifying the expert in question answering community. This approach uses the convolutional neural network and combines user feature representations with question feature representations to compute scores that the user who gets the highest score is the expert on this question. Unlike prior work, this method does not measure expert based on measure answer content quality to identify the expert but only require question sentence and user embedding feature to identify the expert. Remarkably, Our model can be applied to different languages and different domains. The proposed framework is trained on two datasets, The first dataset is Stack Overflow and the second one is Zhihu. The Top-1 accuracy results of our experiments show that our framework outperforms the best baseline framework for expert identification.

\end{quote}
\end{abstract}

\section{Introduction}
Question answering communities, such as Stack Overflow, Yahoo Answers, and Zhihu, are an emerging type of online social network where users can post questions, which are answered by users from the same community. With their growing popularity, QA communities have become more efficient sources of knowledge due to the direct interaction between users with experts in certain areas.

Effectively identifying the expert in each domain is a key to better user engagement. This results in a recommendation system, where given a question, the user that has the most expertise in the question is identified and recommended. In order to achieve this, two essential elements are needed:

\begin{itemize}
\item Understanding the content of the questions.
\item Learning the expertise of each user in each domain.
\end{itemize}

Recent few years, natural language understanding based on deep learning has become a hot topic and made tremendous progress. In this paper, we utilize the convolutional neural network, which is a popular framework for deep learning, to implement the expert identification that is the intractable problem to be solved in recommendation system field. With the dataset, each question has a number of users. In these users, the question only has one expert, and this expert has the highest agree number from users for this question. Our goal has two steps. The first goal is that identify whether its user is expert or not from the existing dataset has both question and answer users. The second goal is that find an expert on a question.

In this paper, we address a CNN model-based expert identification framework, which can combine user feature representations with question feature representations to compute a score, which the expert will get the highest score in this question.

The above definition is general. For each question, we can assume that a lot of users answer this question, but only one of these users is the unique expert. For these one question and users matching, we create a user candidate pool to solve it. In fact, it is not hard to create this candidate pool. The reason is that we already collect a large number of users from the dataset. In the dataset, each question matches a lot of answers, but each answer only matches one user id. Based on these large number of users, we can build this user candidate pool. Then we can randomly choose some users for the question, no matter this user's expertise is related to current question field or not, and then we can identify the unique expert for this question.

Until now, all of the tasks are based on question and answer pairs, we can only choose the best answer for it, but we can not choose the expert for this question. So if a question doesn't have the answer, we can not give the solution to this question. But with question and user pair, we can solve this question. Because we can find the expert who is the best match this question and then invite this user to answer this question. Unfortunately, no papers and no researches are based on the user to solve the question. To our best knowledge, our experiments are also the first time to solve the expert identification problem from question contents and user expertise. So this is the first contribution for this paper.

In this paper, our experiments are base on two datasets. The first dataset is Stack Overflow, and the second one is Zhihu. All of these two datasets consists of four parts: training part, dev part, test1 part and test2 part. One advantage of these two datasets is that we can found the user id. With these user ids, we can easily to use a fashionable method, DeepWalk, to build up a user vector representation. The other advantage of these two datasets is that, unlike the data release from IBM paper, the data domain is only focused on insurance domain. But for our datasets, the Stack Overflow has totally 10 domains, such as AI domain, Apple domain, and AskUbuntu domain, etc. For the Zhihu dataset, we totally have 100 different domains, so this is also the second contribution for this paper.

The rest of the papers is organized as follows: Sec.2 provides the model description, such as DeepWalk, Word2Vec, Glove and convolutional neural network; Sec.3 shows the experiment details; Sec.4 describes the results for the experiments; Sec.5 is the related work and Sec.6 we describe the conclusion.

\section{Model Description}
In this section, we describe the proposed deep learning framework. 
Firstly, we learn a distributed vector representation of a given question and then we extract the user vector which was trained from DeepWalk.
Finally, we compute a score to measure the matching degree and choose the highest one as the expert.

\subsection{DeepWalk}
DeepWalk is a novel approach for learning latent representations of vertices in a network(\citeauthor{perozzi2014deepwalk}). These latent representations encode social relations in a continuous vector space, which is easily exploited by statistical models. DeepWalk generalizes recent advancements in language modeling and unsupervised feature learning (or deep learning) from sequences of words to graphs. DeepWalk uses local information obtained from truncated random walks to learn latent representations by treating walks as the equivalent of sentences. We demonstrate DeepWalk's latent representations on several multi-label network classification tasks for social networks such as BlogCatalog, Flickr, and YouTube. Our results show that DeepWalk outperforms challenging baselines which are allowed a global view of the network, especially in the presence of missing information. DeepWalk representations can provide F1 scores up to 10 percent higher than competing methods when labeled data is sparse. In some experiments, DeepWalk representations are able to outperform all baseline methods while using 60 percent less training data. DeepWalk is also scalable. It is an online learning algorithm which builds useful incremental results and is trivially parallelizable. These qualities make it suitable for a broad class of real-world applications such as network classification, and anomaly detection. 

DeepWalk is an approach recently proposed for social network embedding, which is only applicable for networks with binary edges.   For each vertex, truncated random walks starting from the vertex are used to obtain the contextual information, and therefore only second-order proximity is utilized.

We propose to use a deep learning method, DeepWalk, to learn user representations of a graph's vertices. DeepWalk is a classic algorithm, the Word2Vec for graphs and for embedding nodes, which was the first time successful generating dimensional representations from natural language processing into social network analysis. User representations are latent features of the dictionary of users which can be useful capture the similarity of each neighborhood.

In terms of vertex representation modeling, this yields the optimization problem:

\begin{equation}
min_{\Phi} -logPr({v_{i-w},...,v_{i-1},v_{i+1},...,v_{i+w}} | {\Phi}(v_i))
\end{equation}

As in any language modeling algorithm, the only required input is a corpus and a vocabulary V. DeepWalk considers a set of short truncated random walks its own corpus and the graph vertices as its own vocabulary (V = V ). While it is beneficial to know the V and the frequency distribution of vertices in the random walks ahead of the training, it is not necessary for the algorithm to work as we will show in 4.2.2.

\subsection{Word Embedding}
As is well known, word embedding based CNN architecture has been a hot topic in recent few years. All of this models use pre-trained word vectors as input, and these word vectors have the fixed length. Johnson and Zhang created a CNN model which use one-hot vector representations as input. In this part, we separately use Word2Vec models(\cite{mikolov2013distributed}) and Glove models(\cite{pennington2014glove})  to train the word embeddings before we trained CNN model. Word2Vec and Glove are two efficient and effective models for learning high-quality distributed vector representations that capture a large number of precise syntactic and semantic word relationships.

Word2Vec provides an efficient implementation of the CBOW architecture and skip-gram architectures for computing vector representations of words. CBOW uses the context given by a local window to predict a (known) center word, and skip‐gram works exactly the other way around, using a given center word to predict it’s context.  These representations can be subsequently used in many natural language processing applications and for further research.


Glove is an unsupervised learning algorithm for obtaining vector representations for words. Training is performed on aggregated global word-word co-occurrence statistics from a corpus, and the resulting representations showcase interesting linear substructures of the word vector space.

In machine learning, the words in the sentences are often represented as Bag of Words and the advantage of BoW is that its method can replace the arbitrary length of words with the fixed length vectors. However, Bag of words doesn't classify the word which in the same context. For example, words "car" and "automobile" are often used in the same context.  Bag of words also ignores the context of words. The problem becomes more serious refer to sentences. For example, “Buy used cars” and “Purchase old automobiles” has the same meaning. Fortunately, a large number of works are the focus on learning important representations from text documents, so that is the reason we use Word2Vec or GloVe models to generate the word vector embedding.

With the Zhihu dataset, the most difficult is that the characters are Chinese in the question part. Unlike English word can be immediately training as high-quality distributed vector representations, the Chinese characters need to use word segmentation algorithm to split sentence at first, then put the output into the Word2Vec algorithm and Glove algorithm separately.

\subsection{Convolutional neural network}
In this paper, the question part of the Q-USER-CNN is based on Convolutional Neural Network. So we start from the question sentence which we convert to a sentencing matrix,$\mathbb{R}^{m \mathrm{x} n}$, which m is the row number for the sentence length and which n is the column number for word vector representations.  After generating the question token matrix, we can think this matrix as a 2D image\citeauthor{zhang2015sensitivity}, then effectively use different size of filters to operate convolution and max-pooling. We put four filter region sizes: 2, 3, 4 and 5, each of which has 500 filters. 

As is well-known, the widespread and frequent method to compute the activation function f is sigmoid function with $ f(x) = (1 + e^{-x})^{-1} $ and tanh function with $ f(x) = tanh(x) $. However, the drawback of sigmoid activation function is that the result of the sigmoid function is always positive, during a given step of gradient descent, the weights will either move to the positive direction or the negative one, and the value always becomes fixed. The disadvantage of tanh activation function is that the speed is slow. Luckily, \citeauthor{nair2010rectified} generate a non-saturating nonlinearity named Rectified Linear Units (ReLUs) with $ f(x) = max(0,x) $ to solve this problem. With ReLUs activation function, CNN model treating ReLUs as activation function can train several times faster than use tanh activation function(\citeauthor{krizhevsky2012imagenet}).  So in this paper, we replace the non-saturating nonlinearity, ReLUs, with saturating nonlinearities.

Both question sentence length in Stack Overflow dataset and in Zhihu dataset is fixed 50 words. Especially, the dimensionality of the vector generated by different filters in the convolutional layer will change based on filter region size.
\begin{equation}
d = n - m + 1
\end{equation}
where d is the dimensionality of the result vector, n is the number of sentence length, and m is the filter region size. 

For example, When the filter region size is 2, we will generate a vector which the dimension is 49 after the convolutional layer's operation. It is worth to point out that each size of filter region will generate 500 different vectors. The reason to generate this large number of vectors is that it can spontaneously learn complementary features. Then we put these vectors generated from convolutional layer as inputs into the max-pooling layer. The most common strategy is named 1-max-pooling which is address from \citeauthor{boureau2010theoretical}. In this layer, each vector extracts one scalar and merge all the scalars together to generate a new vector as output.

In order to decrease dimension from the output of the max-pooling operation to the same dimension of user vector, we add neural network layer to decrease dimension. Then we can generate the vector which has the same dimensionality of the user vector.

\begin{figure*}[h!]
  \centering
  \includegraphics[width=0.75\textwidth]{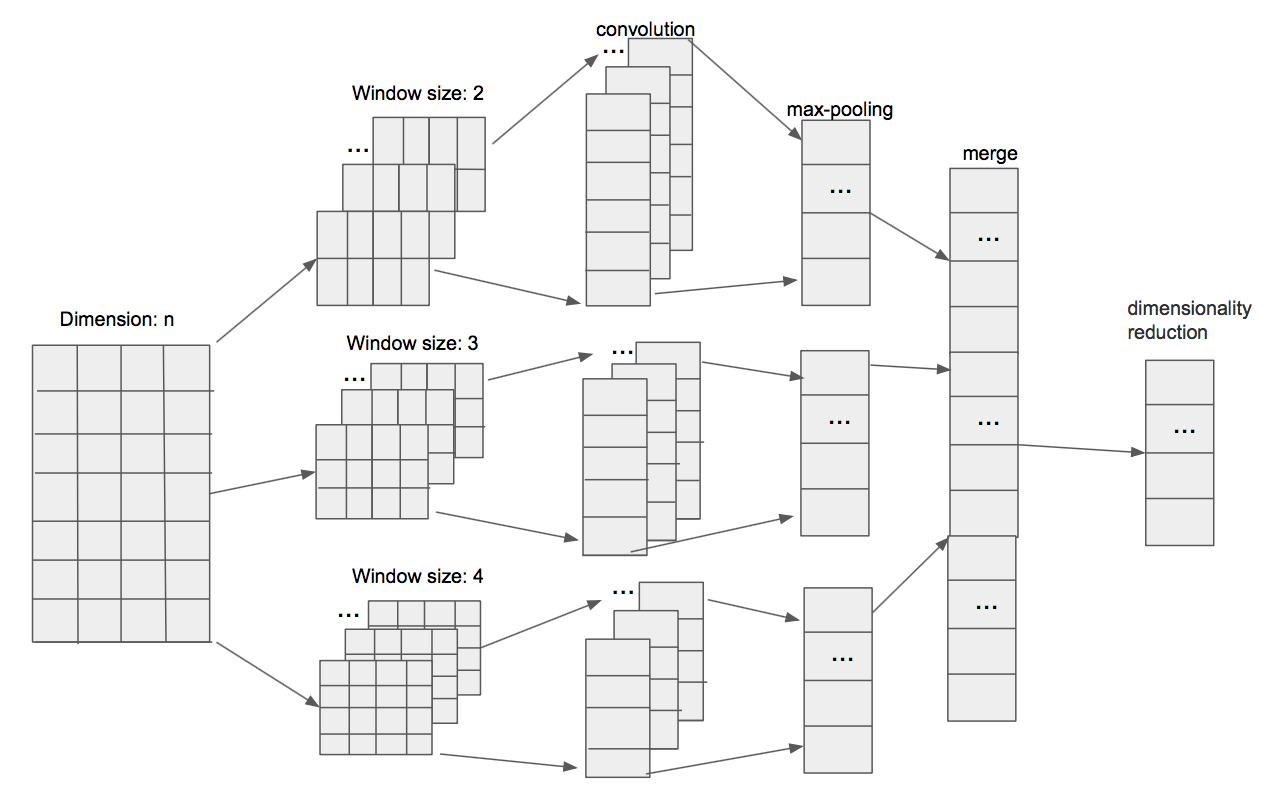}
   \caption{This figure shows the process of the Convolutional neural network for generating the 200 dimension vector. In this figure, the filter window size( region sizes) are separately 2, 3, 4. each filter window has n filters. Firstly, each filter generate the feature maps based on convolutional layer operation. Secondly, we take all of this feature maps into max pooling layer, in the other word, we choose the largest number in each feature map. Then we can generate 3 vectors which of each vector has n dimension. Thirdly we merge these three vectors together as the new feature vector to represent the sentence content. Finally, we decrease the dimensions for this new feature vector which the dimension is same as the user vector representation. }
\end{figure*}

\subsection{Training and Loss function}
we train a deep learning model to learn all the desired modules from data, with minimum to no human interference. To do so, we solve the following optimization problem:

\begin{equation}
\resizebox{.9\hsize}{!}{$ min\sum_{u^+ ,u^- ,q}max[0, margin - (cos(v_{u^+}, h(q, \theta)) - cos(v_{u^-}, h(q, \theta)))] $}
\end{equation}

where q stands a question, cos means cosine similarity, $u^+$,$u^−$ denotes two users such that $u^+$ has higher cosine similarity (number of upvotes, thumbs up, agree numbers, etc.) than $u^−$, i.e., $cos(u^+, q) > cos(u^-, q)$. $ v_u \in R^d $ is a vector representation of a user u, which is pre-trained with the DeepWalk method. $h(q; \theta) \in R^d$ is a function that converts a question q into a d-dimensional vector. While in theory h can be a broad family of functions, we let h be implemented by a deep model and $\theta$ is the learning parameter. In this paper, we use CNN model. The overall goal of the above optimization problem is thus to learn $v_u$ and h, given the contents of questions and the vector representations of each user gets for answering each question. All the parameters can be learned end-to-end, with larger amounts of data available.
Once trained, given a question q, one can easily identify the expert user by:
\begin{equation}
userid = argmax_u(cos(v_{u}, h(q, \theta)) 
\end{equation}
where q means the question content, $u$ describes the userid, cos means cosine similarity.
 
\subsection{Architectures}
In this subsection, we describe the overall architecture of our Q-USER-CNN framework. Figure 2 shows the Architecture of Q-UEER-CNN. 

For the question part, the input is the question sentence. The length of the question sentence is fixed 50. Before we put the input into CNN model, we need to transform each word into word vector representation. We separately used Word2Vec, glove as pre-training word vector representation. After each word in the question sentence transform to the word vector, it provided us a text matrix $\mathbb{R}^{m \mathrm{x} n}$ as an input, which m is the number of tokens in one sentence and n is the size of the dimension for each token. In our task, m is 50 and n are 100. As we all know, all the question sentences have an inherent sequential structure for itself. The pooling in the Fig.x is the max pooling layer. Then we merge all the max-pooling vector together to generate the Merge layer. We put the Merge layer into the full connection layer to generate a 200 dimension vector as output. 

For the user id part, the input is the user id. We already used DeepWalk method to pre-train all the user id as 200 dimensions user id vector representation. So we only need to lookup the user index table and extract the matching vector representation for this user. 

Finally, we generate the question part vector and the user id part dimension vector, we can generate the score, which is the cosine similarity between these two vectors. The result of Q-USER-CNN with Word2Vec and Q-USER-CNN with Glove are shown in table x.

\begin{figure}[h!]
  \centering
  \includegraphics[width=0.5\textwidth]{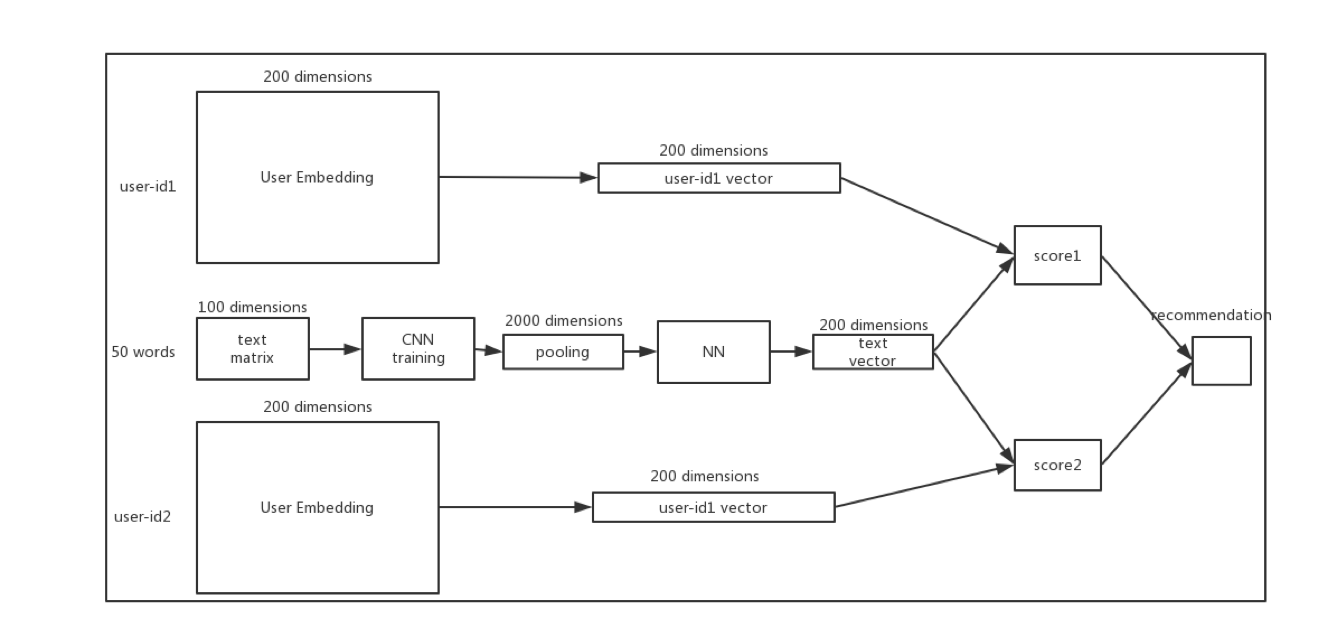}
  \caption{Q-USER-CNN}
\end{figure}

\begin{figure}[h!]
  \centering
  \includegraphics[width=0.5\textwidth]{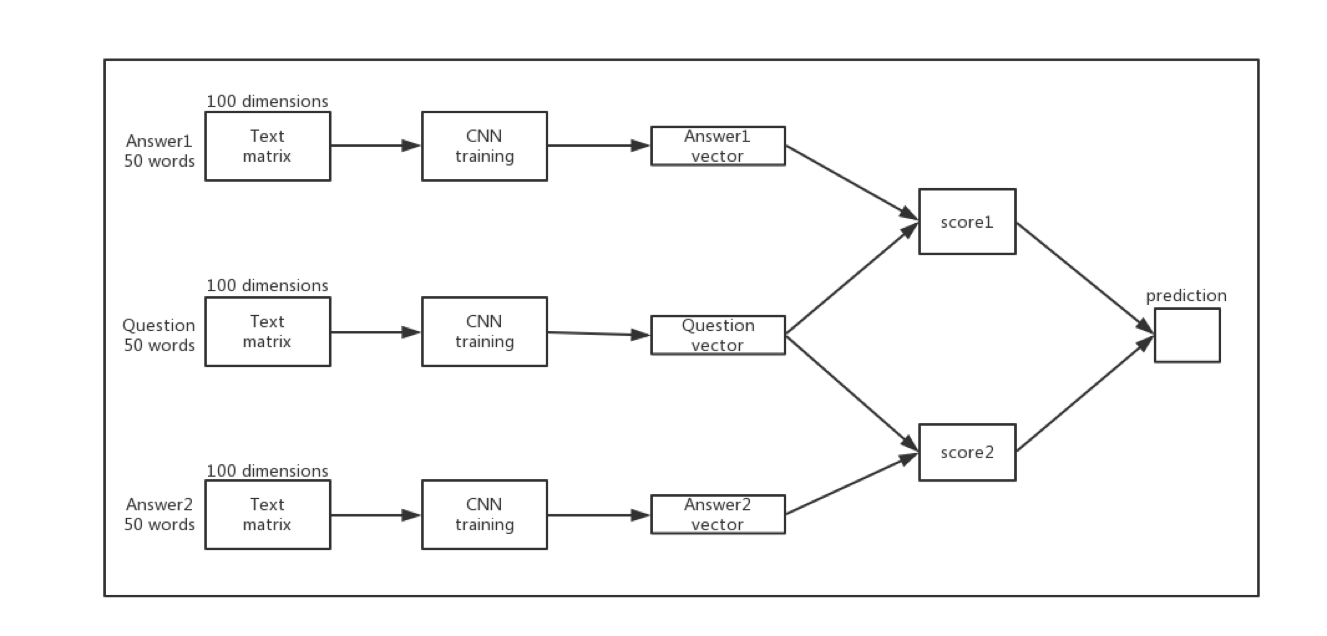}
  \caption{Q-A-CNN}
\end{figure}

\section{Experiments}
\subsection{Baseline}
Our baseline is used question-answer pair and then choose the best answer which got the best score on this question and then choose the user who answer this question(\citeauthor{feng2015applying}). The model description is shown in Figure 3. 

In the training part, the input has three part: Question Q, positive Answer1 $A^+$ and negative Answer2 $A^-$. All of the sentence length for this three input is 50 words. Before putting this three input into CNN model, all the words need to transform into word representation. We also separately use Word2Vec and glove models to generate the word embedding at first. Then we put the text matrix into CNN model and 1-max pooling. Finally, we separately compute the $cos(A^+, q)$ and $cos(A^-, q)$. the loss function is shown below:

\begin{equation}
\resizebox{.9\hsize}{!}{$ max[0, margin - (cos(v_{A^+},v_{Q}) - cos(v_{A^-},v_{Q}))] $}
\end{equation}

In the testing part, given the question and candidate answers, we only need to compute the cosine similarity. After we compute all the cosine similarity, the largest score is the best answer to the current question. The equation is as follow:

\begin{equation}
best answer = argmax_u(cos(v_{A_i},v(Q) ) 
\end{equation}
which i is the ith candidate answer

Unfortunately, the downside of this method is obvious. As is well known, the input of CNN model need fixed length words, but the length of question and answer is not fixed length. For the question part and answer part in QA pairs, an inevitable challenge is that on the one hand the length of an answer may very large, maybe 200 words or even more. But on the other hand, the length of the question is at most 50 words. It is impractical to easily truncate the answer words and use the first 50 words because the contents which truncate are also important.

The second inevitable challenge is that each work in the sentence needs to transform from words to vector representation. So each word needs to look up the table and finds the matching vector representation for current words. If the answer sentence is fixed 50 words, we need to lookup table 50 times. Unfortunately, with the Zhihu dataset, we totally have 970,000 training data, and the index table of word vector has 110,694 words. But with the user id representation, the length of user id is fixed only 1, it can be saved 50 times for each question and answer pair. Furthermore, the index table of user representation also much less than the index table of word representation, it only has 94399 users. So with this two advantage of Q-USER-CNN, the training process can save a lot of times, and we even don't need large hardware support.

\subsection{Datasets}
We experimented our Q-USER-CNN method based on two datasets. Both of these two datasets from the Question Answering forum. The first one selected from StackOverflow Question Answering forums, the second one selected from Zhihu Question Answering Community. In the following, we will briefly introduce these datasets.

StackOverflow is one of the most famous big Question Answering forum which refers to computer science development questions. In this Question Answering forum, there is an ocean of questions and answers posted within categories, which include a wide variety of fields. All the users can conveniently ask their technical questions and post to StackOverflow and then would receive answer feedback effectively and efficiently there. If already existing the close questions, the users can quickly and immediately find the existing answers. With different answers for the same question, guests can easy to get the best answer either chosen by the user who asked the question or selected by other users which the answer received the largest number of votes. We split this collection into two part of datasets. These two datasets have the same questions. The first dataset use for Q-USER-CNN model. The second one use for Q-A-CNN model. Both of this two dataset with 90,000 training and 10,000 test cases.

We collected a large dataset from zhihu.com, which is the largest QA community in Chinese. We will release the dataset for public research. The reason we use this dataset is that not only this dataset has different domains, but also the language of this dataset is Chinese. In this dataset, each question has the matching user id and the matching answer. And for each answer has an agree number which is the source from the other user who views this question and feels this question is useful. So when one user usually answers the question for one domain and receive a lot of agree number, it can obviously justify this user is the expert in the specific domain. We also split this collection into two part of datasets. These two datasets have the same questions. The first dataset use for Q-USER-CNN model. The second one use for Q-A-CNN model. The dataset information is shown in table 1, the pre-trained vector representation is shown in table2. 

\begin{table}[h] 

\begin{center}
\begin{tabular}{|c|c|c|}
    \hline
    Dataset & Set  & QU/QA pair\\  
    \hline
    Stack Overflow & Train   & 97,000\\
           & Dev    & 5000\\
           & Test1  & 9,700\\
           & Test2  & 9,700\\
    \hline
    Zhihu  & Train & 1,200,000\\
           & Dev  &  2,500\\
           & Test1  &  10,000\\
           & Test2  &  10,000\\
    \hline
\end{tabular}
\end{center}
\caption{Stack Overflow and Zhihu dataset}
\end{table}

\begin{table}[h] 
\begin{center}
\begin{tabular}{|c|ccc|}
    \hline
    Dataset &  Vector type & count & dimension\\  
    \hline
    Stack Overflow & question  & 164217 & 100\\
                   & userid    & 53134  & 200\\
    \hline
    Zhihu & question  & 110694 & 100\\
                   & userid    & 94399  & 200\\
    \hline
\end{tabular}
\end{center}
\caption{vector statistic}
\end{table}

For each line of the first dataset, we composed the question, user1, and user2. The performance of user1 is always better than user2 because in zhihu.com, each answer refers to this question has an agree number. The agree number of user1 who answered this question is always better than user2.

For each line of the second dataset, we composed the question, answer1, and answer2. For the same reason, The performance of answer1 is always better than answer2.

\subsection{Hyperparameters}

In this paper, we built up this deep learning framework by using Pytorch. The word embedding in question part (100 dimensions) is trained by Word2Vec and by Glove and used for initialization, the user id embedding in user feature part (200 dimensions) is trained by DeepWalk. Especially, word embeddings in question part are also parameters and are optimized for the Q-USER-CNN framework. And the optimization strategy separately used Adam optimizer(\citeauthor{kingma2014adam}) and Stochastic Gradient Descent optimizer (\citeauthor{bottou2010large}). However, SGD methods have many disadvantages. One key disadvantage of SGDs is that they require much manual tuning of optimization parameters such as learning rates and convergence criteria(\cite{ngiam2011optimization}). Remarkably, the learning rate is 0.00001(and we also use the other learning rate, but 0.00001 got the best performance), the dropout(\cite{srivastava2014dropout} \citeauthor{ba2013adaptive}) is 0.5, and the margin in loss function is 0.1. With the hardware environment, we use K40Ti as GPU.

-

\section{Results}
The results on ZhiHu dataset are summarized in Table 3 and Table 4.
For the hyperparameter column part, the first one is word embedding dimension, the second one is user embedding dimension, the third one is learning rate. As is shown in the tables, with the Stack Overflow dataset, it got the best performance when region sizes are (2, 3, 4, 5), the learning rate is 1e-4, the word embedding chose Glove method, the optimizer selected Adam in the Q-USER-CNN model. Simultaneously, with the Zhihu dataset, it got the best performance when region sizes are (3, 4, 5), the learning rate is 1e-5, the word embedding chose Glove method, the optimizer selected Adam in the Q-USER-CNN model. Especially, the top1 accuracy with Q-USER-CNN can outperform Q-A-CNN model in the Stack Overflow dataset.

\begin{table*}[h] 
\begin{center}
\begin{tabular}{|c|c|c|c|c|c|c|}
    \hline
    Method & region sizes & hyperparameter & word embedding & optimizer & test1 Top-1 & test2 Top-1\\  
    \hline
    Q-USER-CNN 
           & (2,3,4) &100 200 1e-4 & Word2Vec & SGD & 78.77 & 84.26   \\
           & (2,3,4) &100 200 1e-5 & Word2Vec & Adam & 79.69 & 84.90   \\
           & (3,4,5) &100 200 1e-4 & Word2Vec & SGD & 78.79 & 84.41 \\
           & (3,4,5) &100 200 1e-5 & Word2Vec & Adam & 79.65 & 84.96 \\
           & (2,3,4,5) &100 200 1e-4 & Word2Vec & Adam & 79.92 & 85.24 \\
           & (2,3,4,5) &100 200 1e-5 & Word2Vec & Adam & 80.04 & 85.28 \\
           & (2,3,4) &100 200 1e-4 & Glove & SGD & 78.80 & 84.30   \\
           & (2,3,4) &100 200 1e-5 & Glove & Adam & 79.71 & 84.90   \\
           & (3,4,5) &100 200 1e-4 & Glove & SGD & 78.79 & 84.42 \\
           & (3,4,5) &100 200 1e-5 & Glove & Adam & 79.70 & 84.98 \\
           & (2,3,4,5) &100 200 1e-4 & Glove & Adam & 79.92 & 85.26 \\
           & (2,3,4,5) &100 200 1e-5 & Glove & Adam & 80.06 & 85.32 \\
    \hline
    Q-A-CNN    
           & (2,3,4) &100 200 1e-4 & Word2Vec & SGD & 77.22 & 82.81   \\
           & (2,3,4) &100 200 1e-5 & Word2Vec & Adam & 78.10 & 83.14   \\
           & (3,4,5) &100 200 1e-4 & Word2Vec & SGD & 77.24 & 82.81 \\
           & (3,4,5) &100 200 1e-5 & Word2Vec & Adam & 78.13 & 83.20 \\
           & (2,3,4,5) &100 200 1e-4 & Word2Vec & Adam & 78.69 & 83.88 \\
           & (2,3,4,5) &100 200 1e-5 & Word2Vec & Adam & 78.72 & 83.91 \\
           & (2,3,4) &100 200 1e-4 & Glove & SGD & 77.27 & 82.83   \\
           & (2,3,4) &100 200 1e-5 & Glove & Adam & 78.15 & 83.18   \\
           & (3,4,5) &100 200 1e-4 & Glove & SGD & 77.24 & 82.88 \\
           & (3,4,5) &100 200 1e-5 & Glove & Adam & 78.16 & 83.22 \\
           & (2,3,4,5) &100 200 1e-4 & Glove & Adam & 78.71 & 83.91 \\
           & (2,3,4,5) &100 200 1e-5 & Glove & Adam & 78.76 & 83.97 \\
    \hline
\end{tabular}
\end{center}
\caption{Stack Overflow Result}
\end{table*}

\section{Related work}

As is well known, expert identification approaches can be roughly divided into two types. The first type is feature-based approaches, the other type is graph-based approaches. With the graph-based approach, the dominant algorithms are PageRank, HITS, ExpertRank, etc. With the feature based feature-based approaches, the obvious operation is that generate user embedding in a high dimensional space, such as DeepWalk, Word2Vec, or Glove, and then identify expert based on several popular learning methods.

The goal of graph-based approaches in expert identification is to build up a graph analysis which interacts with users and to find experts on the different topics. \citeauthor{jurczyk2007discovering} used link analysis of the underlying graph to identify expert based on Yahoo! answer; \citeauthor{zhang2007expertise} showed us the expertise Ranking algorithms and this method beat most of the graph algorithms which even more complex.

\citeauthor{bouguessa2008identifying} proposed a method based on the number of best answers to build up the user expertizes. In this method, it would find the number of users as experts automatically based on the expertise of users in different answers.

\section{Conclusions}

In this paper, we propose a Q-USER-CNN model to solve the expert identification problems in recommendation system. More importantly, our framework can use in different languages, such as English, Chinese, and can adapt to different Question Answering communities. Based on our work, we address a good way to solve expert identification problem by deep learning method. For the framework which we proposed, we found that it not only outperform previous works but also our framework does not rely on any linguistic tools and our framework can be applied to different domains.

In the future, we would like to further evaluate the models presented in this paper for different tasks, such as answer quality prediction in Community QA, recognizing textual entailment, and machine comprehension of text.

\bibliographystyle{aaai}
\bibliography{bibfile}

\begin{thebibliography}{}

\bibitem[\protect\citeauthoryear{Ba and Frey}{2013}]{ba2013adaptive}
Ba, J., and Frey, B.
\newblock 2013.
\newblock Adaptive dropout for training deep neural networks.
\newblock In {\em Advances in Neural Information Processing Systems},
  3084--3092.

\bibitem[\protect\citeauthoryear{Bottou}{2010}]{bottou2010large}
Bottou, L.
\newblock 2010.
\newblock Large-scale machine learning with stochastic gradient descent.
\newblock In {\em Proceedings of COMPSTAT'2010}. Springer.
\newblock  177--186.

\bibitem[\protect\citeauthoryear{Bouguessa, Dumoulin, and
  Wang}{2008}]{bouguessa2008identifying}
Bouguessa, M.; Dumoulin, B.; and Wang, S.
\newblock 2008.
\newblock Identifying authoritative actors in question-answering forums: the
  case of yahoo! answers.
\newblock In {\em Proceedings of the 14th ACM SIGKDD international conference
  on Knowledge discovery and data mining},  866--874.
\newblock ACM.

\bibitem[\protect\citeauthoryear{Boureau, Ponce, and
  LeCun}{2010}]{boureau2010theoretical}
Boureau, Y.-L.; Ponce, J.; and LeCun, Y.
\newblock 2010.
\newblock A theoretical analysis of feature pooling in visual recognition.
\newblock In {\em Proceedings of the 27th international conference on machine
  learning (ICML-10)},  111--118.

\bibitem[\protect\citeauthoryear{Feng \bgroup et al\mbox.\egroup
  }{2015}]{feng2015applying}
Feng, M.; Xiang, B.; Glass, M.~R.; Wang, L.; and Zhou, B.
\newblock 2015.
\newblock Applying deep learning to answer selection: A study and an open task.
\newblock In {\em Automatic Speech Recognition and Understanding (ASRU), 2015
  IEEE Workshop on},  813--820.
\newblock IEEE.

\bibitem[\protect\citeauthoryear{Jurczyk and
  Agichtein}{2007}]{jurczyk2007discovering}
Jurczyk, P., and Agichtein, E.
\newblock 2007.
\newblock Discovering authorities in question answer communities by using link
  analysis.
\newblock In {\em Proceedings of the sixteenth ACM conference on Conference on
  information and knowledge management},  919--922.
\newblock ACM.

\bibitem[\protect\citeauthoryear{Kingma and Ba}{2014}]{kingma2014adam}
Kingma, D., and Ba, J.
\newblock 2014.
\newblock Adam: A method for stochastic optimization.
\newblock {\em arXiv preprint arXiv:1412.6980}.

\bibitem[\protect\citeauthoryear{Krizhevsky, Sutskever, and
  Hinton}{2012}]{krizhevsky2012imagenet}
Krizhevsky, A.; Sutskever, I.; and Hinton, G.~E.
\newblock 2012.
\newblock Imagenet classification with deep convolutional neural networks.
\newblock In {\em Advances in neural information processing systems},
  1097--1105.

\bibitem[\protect\citeauthoryear{Mikolov \bgroup et al\mbox.\egroup
  }{2013}]{mikolov2013distributed}
Mikolov, T.; Sutskever, I.; Chen, K.; Corrado, G.~S.; and Dean, J.
\newblock 2013.
\newblock Distributed representations of words and phrases and their
  compositionality.
\newblock In {\em Advances in neural information processing systems},
  3111--3119.

\bibitem[\protect\citeauthoryear{Nair and Hinton}{2010}]{nair2010rectified}
Nair, V., and Hinton, G.~E.
\newblock 2010.
\newblock Rectified linear units improve restricted boltzmann machines.
\newblock In {\em Proceedings of the 27th international conference on machine
  learning (ICML-10)},  807--814.

\bibitem[\protect\citeauthoryear{Ngiam \bgroup et al\mbox.\egroup
  }{2011}]{ngiam2011optimization}
Ngiam, J.; Coates, A.; Lahiri, A.; Prochnow, B.; Le, Q.~V.; and Ng, A.~Y.
\newblock 2011.
\newblock On optimization methods for deep learning.
\newblock In {\em Proceedings of the 28th international conference on machine
  learning (ICML-11)},  265--272.

\bibitem[\protect\citeauthoryear{Pennington, Socher, and
  Manning}{2014}]{pennington2014glove}
Pennington, J.; Socher, R.; and Manning, C.~D.
\newblock 2014.
\newblock Glove: Global vectors for word representation.
\newblock In {\em EMNLP}, volume~14,  1532--1543.

\bibitem[\protect\citeauthoryear{Perozzi, Al-Rfou, and
  Skiena}{2014}]{perozzi2014deepwalk}
Perozzi, B.; Al-Rfou, R.; and Skiena, S.
\newblock 2014.
\newblock Deepwalk: Online learning of social representations.
\newblock In {\em Proceedings of the 20th ACM SIGKDD international conference
  on Knowledge discovery and data mining},  701--710.
\newblock ACM.

\bibitem[\protect\citeauthoryear{Srivastava \bgroup et al\mbox.\egroup
  }{2014}]{srivastava2014dropout}
Srivastava, N.; Hinton, G.~E.; Krizhevsky, A.; Sutskever, I.; and
  Salakhutdinov, R.
\newblock 2014.
\newblock Dropout: a simple way to prevent neural networks from overfitting.
\newblock {\em Journal of machine learning research} 15(1):1929--1958.

\bibitem[\protect\citeauthoryear{Zhang, Ackerman, and
  Adamic}{2007}]{zhang2007expertise}
Zhang, J.; Ackerman, M.~S.; and Adamic, L.
\newblock 2007.
\newblock Expertise networks in online communities: structure and algorithms.
\newblock In {\em Proceedings of the 16th international conference on World
  Wide Web},  221--230.
\newblock ACM.

\bibitem[\protect\citeauthoryear{Zhang and
  Wallace}{2015}]{zhang2015sensitivity}
Zhang, Y., and Wallace, B.
\newblock 2015.
\newblock A sensitivity analysis of (and practitioners' guide to) convolutional
  neural networks for sentence classification.
\newblock {\em arXiv preprint arXiv:1510.03820}.

\end{thebibliography}

\end{document}